\documentclass{article}

\usepackage{amsmath,amsfonts,bm}

\def\eqref#1{equation~\ref{#1}}

\def\1{\bm{1}}

\def\vv{{\bm{v}}}

\def\vy{{\bm{y}}}

\def\mA{{\bm{A}}}

\def\mD{{\bm{D}}}

\def\mL{{\bm{L}}}

\def\mP{{\bm{P}}}

\def\mV{{\bm{V}}}

\DeclareMathAlphabet{\mathsfit}{\encodingdefault}{\sfdefault}{m}{sl}
\SetMathAlphabet{\mathsfit}{bold}{\encodingdefault}{\sfdefault}{bx}{n}

\DeclareMathOperator*{\argmin}{arg\,min}

\usepackage{latexsym}   
\usepackage{amsbsy}
\usepackage{amssymb}
\usepackage{amsmath}
\usepackage{verbatim}
\usepackage[final]{graphicx}
\usepackage{color}
\usepackage{enumitem}
\usepackage{amsthm}
\usepackage{subfigure}

\allowdisplaybreaks

\newtheorem{theorem}{Theorem}

\newcommand{\be}{\begin{equation}}
\newcommand{\ee}{\end{equation}}
\newcommand{\beq}{\begin{equation}}
\newcommand{\eeq}{\end{equation}}
\newcommand{\beqy}{\begin{eqnarray}}
\newcommand{\eeqy}{\end{eqnarray}}
\newcommand{\beqynn}{\begin{eqnarray*}}
\newcommand{\eeqynn}{\end{eqnarray*}}

\newcommand{\ba}{\begin{array}}
\newcommand{\ea}{\end{array}}

\newcommand{\bmx}{\begin{bmatrix}}
\newcommand{\emx}{\end{bmatrix}}
\newcommand{\bsmx}{\left[\begin{smallmatrix}}
\newcommand{\esmx}{\end{smallmatrix}\right]}
\newcommand{\bmxc}[1]{\left[\begin{array}{@{}#1@{}}}
\newcommand{\emxc}{\end{array}\right]}

\newcommand{\bt}[1]{\begin{tabular}{#1}}
\newcommand{\et}{\end{tabular}}

\newcommand{\bc}{\begin{center}}
\newcommand{\ec}{\end{center}}
\newcommand{\ben}{\begin{enumerate}}
\newcommand{\een}{\end{enumerate}}
\newcommand{\bi}{\begin{itemize}}
\newcommand{\ei}{\end{itemize}}

\newcommand{\diag}{\mathrm{diag}}

\newcommand{\A}{\mathbf{A}}

\newcommand{\E}{\mathbf{E}}

\newcommand{\methodfullname}{Shapley Credit Assignment Rewards}
\newcommand{\methodacronym}{SCAR}

\usepackage[utf8]{inputenc} 
\usepackage[T1]{fontenc}    
\usepackage{hyperref}       
\usepackage{url}            
\usepackage{booktabs}       
\usepackage{amsfonts}       
\usepackage{nicefrac}       
\usepackage{microtype}      
\usepackage{xcolor}         
\usepackage{tabularx}

\usepackage{graphicx}       
\usepackage{caption}
\usepackage{subcaption}     
\usepackage{wrapfig}        
\usepackage{adjustbox} 

\usepackage{listings} 
\usepackage{enumitem} 
\usepackage{tcolorbox}
\tcbuselibrary{breakable} 

\lstdefinestyle{mypromptstyle}{
  basicstyle=\ttfamily\footnotesize, 
  breaklines=true,                
  showstringspaces=false,         
  columns=flexible,
  keepspaces=true,
  backgroundcolor=\color{gray!10}, 
  frame=none,                     
  xleftmargin=1em,                
  escapeinside={(*@}{@*)},        
}

\usepackage{natbib}

\definecolor{poscontrib}{HTML}{C0392B} 
\definecolor{negcontrib}{HTML}{1F618D} 
\definecolor{zerocontrib}{HTML}{B0B0B0} 

\newcommand{\hlscore}[3]{
  \begin{tabular}{@{}c@{}}
    {\setlength{\fboxsep}{1pt}\colorbox{#3}{\strut\color{black}#1}} \\
    \texttt{#2}
  \end{tabular}%
}

\newcommand{\plainscore}[2]{
  \begin{tabular}{@{}c@{}}
    #1 \\ 
    \texttt{#2}
  \end{tabular}%
}

\title{Towards better dense rewards in Reinforcement Learning Applications}
\author{Shuyuan Zhang$^{1,2}$}
\date{%
    $^1$School of Computer Science, McGill University\\%
    $^2$Mila\\[2ex]%
}

\begin{document}

\maketitle
\section{Introduction}
Finding meaningful and accurate dense rewards is a fundamental task in the field of reinforcement learning (RL) that enables agents to explore environments more efficiently. In traditional RL settings, agents learn optimal policies through interactions with an environment guided by reward signals. However, when these signals are sparse, delayed, or poorly aligned with the intended task objectives, agents often struggle to learn effectively. Dense reward functions, which provide informative feedback at every step or state transition, offer a potential solution by shaping agent behavior and accelerating learning.

Despite their benefits, poorly crafted reward functions can lead to unintended behaviors, reward hacking, or inefficient exploration. This problem is particularly acute in complex or high-dimensional environments where handcrafted rewards are difficult to specify and validate.

To address this, recent research has explored a variety of approaches, including inverse reinforcement learning, reward modeling from human preferences, and self-supervised learning of intrinsic rewards. While these methods offer promising directions, they often involve trade-offs between generality, scalability, and alignment with human intent. This proposal explores several approaches to dealing with these unsolved problems and enhancing the effectiveness and reliability of dense reward construction in different RL applications.
\paragraph{Proposal Structure} 
\begin{itemize}
    \item Section \ref{sec:Background} introduces the relevant notation and background knowledge. 
    \item In Section \ref{1stwork}, we present Graph-Guided subGoal representation Generation (G4RL), a novel graph encoder-decoder framework for goal-conditioned scenarios. This work's primary objective is to boost the performance of existing Goal-Conditioned Hierarchical Reinforcement Learning (GCHRL) methods. Traditional GCHRL often relies on the Euclidean distance between the current state and the assigned subgoal to generate an intermediate reward for the low-level agent. This reliance can be problematic, as the geometric distance in the state space often fails to align with the true transition distance or accessibility between states. Consequently, this leads to inaccurate and misleading intermediate reward signals. Our G4RL framework addresses this by introducing a method to encode subgoal representations using the underlying transition graph of the environment. By leveraging this graph, our encoder-decoder architecture generates subgoal embeddings that inherently reflect the true transition dynamics. This approach produces more accurate and informative dense rewards, significantly enhancing the effectiveness of typical GCHRL methods.
    \item Section \ref{2ndwork} addresses the challenge of sparse rewards in traditional Reinforcement Learning from Human Feedback (RLHF) and introduces our proposed solution: Shapley Credit Assignment Rewards (SCAR). Vanilla RLHF suffers from reward sparsity because the reward model typically provides a usable signal only upon the completion of a full response. This lack of intermediate feedback hinders efficient learning. While previous methods have attempted to use dense rewards by evaluating partial sequences, they often introduce inaccurate rewards because the partial sequences lie outside the reward model's training distribution (out-of-distribution). Our methodology resolves both issues simultaneously. It leverages the Shapley value from cooperative game theory to assign a statistically sound credit score to each component (e.g., token or phrase) of a full generated sentence. This approach effectively converts the sparse, sentence-level reward into an accurate, dense signal for every part of the sequence, thus promoting more robust and efficient learning. 
    \item In Section \ref{3rdwork}, we outline an ongoing project focused on enhancing agents’ continual learning capabilities. This is achieved by introducing an auxiliary reward signal derived from structural similarities among tasks. The core idea is to guide the agent's learning process by having it identify structural commonalities across different tasks. When the agent encounters a new task whose underlying structure is similar to previous experiences, it leverages this recognized similarity as a crucial learning heuristic and guidance signal, thereby accelerating adaptation and improving performance in the continual learning setting.
\end{itemize}

\section{Background}
\label{sec:Background}
\subsection{Markov Decision Processes}
As the most common framework for modeling reinforcement learning scenarios, the Markov Decision Process (MDP) \citep{puterman2014markov} is introduced as a tuple $<\mathcal{S}, \mathcal{A}, P, R, \gamma>$, defined as follows: At each time step $t$,  the agent observes the current state $s_t \in \mathcal{S}$ provided by the environment and chooses an action $a_t \in \mathcal{A}$ according to its internal policy $\pi(a_t|s_t)$, which specifies the probability of choosing action $a_t$ given state $s_t$. 
The action is then executed, and the interaction with the environment leads the agent to a new state $s_{t+1}$ according to a transition probability function $P(s_{t+1}|s_t, a_t)$ which is known only to the environment.  
Subsequently, the agent receives a reward $r_t$, determined by the reward function $R(s_t,a_t)$ that evaluates the action taken in the current state and is also only visible to the environment. The agent aims to learn an optimal policy $\pi$ to maximize the expected discounted cumulative reward $\mathbb{E}_{\pi} \big[\sum_{t=0}^T \gamma^t r_t \big] $, where $\gamma$  (with $0\leq\gamma<1$) is a pre-defined discount factor used to prioritize immediate rewards over distant future rewards, thereby ensuring that the total reward remains finite.

\subsection{Goal-conditioned Hierarchical RL (GCHRL)}

Goal-Conditioned Reinforcement Learning (GCRL) trains agents to achieve specific goals, which are the target states. The agent receives an additional goal input $g_t$ along with the state input $s_t$ and learns a policy $\pi(a_t|g_t, s_t)$ that aims to achieve this goal. Goals are represented explicitly in the input to the policy, guiding the agent’s actions towards desired outcomes.
The reward function is goal-dependent, providing positive feedback when the agent successfully reaches the desired goal state.

To deal with large and complex environments, 
Goal-Conditioned Hierarchical Reinforcement Learning (GCHRL) \citep{nachum2018data, zhang2022adjacency} decomposes the learning task into a hierarchy of smaller, more manageable sub-tasks. Typically, there are two levels of agents. At time step $t$, the high-level agent chooses a subgoal $g_t$, a representation of a target state, 
and assigns it to the low-level agent to achieve as part of the overall task. 
This choice is made by sampling $g_t$ from the high-level policy $\pi_h(g_t|\phi(s_t))$,  where $\phi: s \mapsto \mathbb{R}^d$ is the state representation function which gives a condensed representation of the state.

Each state $s_t$ can be mapped to its subgoal feature $g(s_t)$ by a subgoal feature extractor. 
Note that $g(s_t)$ is not the same as $g_t$. 
The former, $g(s_t)$, is the learned subgoal feature of the current state $s_t$, while the latter, $g_t$, is the target state we aim to reach from the state $s_t$ in one step or multiple steps.

Given the subgoal $g_t$ sampled from 
the high-level policy  $\pi_h(g_t|\phi(s_t))$ for the current time step $t$ and the state representation vector $\phi(s_t)$, a low-level agent executes action $a_t$ based on the low-level policy $\pi_l(a_t|\phi(s_t), g_t)$. The low-level agent is trained using the intrinsic (dense) reward signal $r_{\text{int}}(s_t, g_t, a_t, s_{t+1})=-\|\phi(s_{t+1})-g_t\|^2_2$ to encourage it to achieve the subgoal.

Both agents can be implemented by any policy-based methods, including those introduced in previous works on policy gradients such as \cite{fujimoto2018addressing, haarnoja2018soft} and \cite{schulman2017proximal}.

\subsection{RL for Text Generation}
Reinforcement Learning (RL) has been increasingly leveraged for fine-tuning LLMs in text generation tasks \cite{ryang-abekawa-2012-framework, NIPS2016_2f885d0f, li-etal-2016-deep, buck2018ask, bahdanau2017an}. Unlike standard supervised fine-tuning which relies on maximizing the likelihood of ground-truth sequences, RL enables optimization directly towards sequence-level objectives or metrics that are not differentiable, such as ROUGE scores in summarization or human preferences for dialogue quality \citep{NEURIPS2020_1f89885d, NEURIPS2022_b1efde53}. A particularly successful application of this is Reinforcement Learning from Human Feedback (RLHF) \cite{NEURIPS2022_b1efde53, NIPS2017_d5e2c0ad, NEURIPS2020_1f89885d, bai2022training}, which has become a widely adopted technique for aligning LLMs with complex human values and instructions. The typical RLHF process involves learning a reward model (RM) from a dataset of human comparisons between different model outputs, followed by fine-tuning the LLM policy to maximize the scalar reward assigned by the RM to the generated text sequences. However, a widely acknowledged challenge in this standard RLHF framework is the inherent sparsity of the reward signal: the RM typically provides feedback only after the entire sequence has been generated \cite{NEURIPS2020_1f89885d, bai2022training}. This terminal reward makes the temporal credit assignment problem—identifying which specific token choices (actions) contributed positively or negatively to the final outcome—particularly difficult \cite{sutton1984temporal}. This difficulty has spurred research into methods for creating denser and more informative reward signals.

\section{Using Spatial Information as Dense Reward in Goal-Conditioned Hierarchical Reinforcement Learning}
\label{1stwork}
This section presents our first published work, Graph-Guided subGoal representation Generation (G4RL) \citep{zhang2025incorporating}.
This method reshapes the subgoal space by utilizing a state graph to incorporate the relative spatial information of visited states, thereby generating better dense rewards for different levels of agents in goal-conditioned RL.

One drawback of previous hierarchical reinforcement learning algorithms \citep{nachum2018data,kim2021landmark,zhang2022adjacency,luo2024goal} 
is that the Euclidean distance  
calculated in the original state representation space between the current state and the intended goal does not accurately reflect the true progress of the low-level agent, as there is rarely a straight path between the current state and the subgoal in the space. 

As a result, the low-level agent trained with such information may receive an inaccurate reward signal, thus impairing its performance. 
Another issue is that, without appropriate constraints,
the high-level agent may propose subgoals that are too difficult to reach, 
wasting exploration steps on pursuing infeasible targets \citep{zhang2022adjacency}. 
The proposed method aims to mitigate both problems by calculating the distance in a subgoal representation space between subgoal representations given by a graph encoder-decoder. 
This graph encoder-decoder captures the actual connectivity between states, ensuring that the generated subgoal representations respect adjacency information.

\subsection{State graph}
To record the visited states and their connections, We maintain a state graph $\mathcal{G}=(\mathcal{V},\mathcal{E})$ with a fixed number $N$ of nodes\footnote{The number of training 
states for the graph encoder-decoder grows quadratically with $N$ because the adjacency weight matrix has $N^2$ elements. The choice of $N$ depends on the machine's capabilities.}. This graph is built and updated during training, with no pre-training using expert data or handcrafted process involved in its construction.

Each node is labelled by the corresponding state and for each node $s_t$, the corresponding state representation vector $\phi(s_t)$, which is also referred to as the node feature, 
is stored.  Edges in the graph represent connectivity between states. The graph is constantly updated during exploration.

We choose the state graph to be undirected for mainly two reasons: \textbf{(1) Efficiency:} An undirected graph requires fewer resources for storage and computation compared to a directed graph, and \textbf{(2) Method compatibility:} The method relies on defining a deterministic distance between state/subgoal representations for each node pair. This is straightforward in an undirected graph but becomes problematic in a directed setting, where distances can be asymmetric or undefined. 

This choice is based on the assumption that G4RL is designed for environments with symmetrical and reversible dynamics. However, some of the experiments demonstrate that G4RL can still enhance performance in partially asymmetric environments; please refer to the experiments section for further details.

\subsubsection{Graph construction}\label{sec:construction}
The graph is initialized with $N$ empty nodes and no edges. The corresponding weighted adjacency matrix $\mA$ is set to an $N\times N$ zero matrix. 
We perform the GCHRL exploration process using randomly initialized policy $\pi_h$ and $\pi_l$. Once the agent encounters a state representation never seen before, that is, the representation is different from any state representations stored in the graph,
as described in equation (\ref{eq:newnode}), it stores the state representation $\phi(s_t)$ as the node feature of an empty node
in the graph and build an edge between this node and the node corresponds to the previous state:
\begin{equation} \label{eq:newnode}
\forall_{s_v \in \mathcal{V}},   \|\phi(s_t)-\phi(s_v)\|_2>\epsilon_d,
\end{equation}
\begin{equation} \label{eq:newnodeedge}
\mA_{\phi(s_t), \phi(s_{t-1})}=\mA_{\phi(s_{t-1}), \phi(s_t)}=1,
\end{equation}
where 
$\epsilon_d$ is a hyper-parameter controlling the distance threshold between state representations.

When the agent encounters a state $s_t$ with feature $\phi(s_t)$ that is similar to several node representations already stored in the graph, it finds the state whose representation is
the closest to the current state feature:

\begin{equation} \label{eq:oldnode}
s_v = \argmin_{s_u: \|\phi(s_t)-\phi(s_u)\|_2 \leq \epsilon_d} \|\phi(s_t)-\phi(s_u)\|_2.
\end{equation}
Then the node $s_v$ is relabeled as $s_t$ and the weight for the edge $(s_{t-1},s_t)$
is updated as follows:
\begin{equation} \label{eq:updateedge}
\mA_{\phi(s_{t-1}), \phi(s_t)}=\mA_{\phi(s_t), \phi(s_{t-1})}:=\mA_{\phi(s_{t-1}),\phi(s_t)}+1.
\end{equation}
Note that a large weight indicates more frequent transitions between the underlying states.

We have used the Euclidean norm to define the distance between feature vectors.  Since some elements may contain more spatial information than others, 
one can use a weighted Euclidean norm to define the distance between state representations instead. 

\subsubsection{Graph updating}
The graph has a fixed number of nodes.
Suppose the graph is now full.
When a new state $s_t$ is encountered, 
if $s_v$ from equation (\ref{eq:oldnode}) exists, as before we relabel the node as $s_t$ 
and perform edge update as shown in equation (\ref{eq:updateedge});
Otherwise, we replace the oldest state node in the graph with the current state node,
delete all edges previously linked to that node,
and create an edge $(\phi(s_{t-1}),\phi(s_t))$ with weight $\A_{\phi(s_{t-1}),\phi(s_t)}=\A_{\phi(s_t),\phi(s_{t-1})}=1$.
Alternatively, we could replace the state node that is most weakly connected to the other nodes--that is, the node with the lowest sum of edge weights.

\subsection{Graph encoder-decoder}
\label{sec:GED}
To enable the assignment of suitable subgoal representations to every possible state, including unseen ones, we use node representations and edges to train a graph encoder-decoder. The parameter updates of the graph encoder-decoder and the policies during policy training
are performed alternately in each episode. 

The encoder-decoder starts training after the graph is full and continues periodically after processing a few trajectories. Section 3.3 will show the details of the training schedule.

The encoder $\E$  maps every state representation $\phi(s)$ to a subgoal representation $g(s)$. 
We use a feed-forward network (FFN) 
with several layers as the encoder $\mathbf{E}$:
\begin{equation} \label{eq:graphe}
g(s)=\mathbf{E}(\phi(s))=\text{FFN}(\phi(s)).
\end{equation}
The weight parameters of the feed-forward network will be learned through training.
The decoder $\mathbf{D}$ takes two subgoal representations as input
and outputs the inner product of these two representations:
\begin{equation} \label{eq:graphd}
\mathbf{D}(g(s_u),g(s_v)) = g(s_u)^Tg(s_v).
\end{equation}
We choose dot-product similarity based on the assumption that the similarity between two nodes, such as the overlap in their local neighbourhoods, is well captured by the dot product of their embeddings. This assumption is supported by prior work in the graph embedding literature \citep{ahmed2013distributed, cao2015grarep, ou2016asymmetric}.

The aim is to use the encoder-decoder structure to predict node relations. 
Naturally we can use $\mA_{\phi(s_u), \phi(s_v)}$ as a measure of the relation 
between nodes $\phi(s_u)$ and $\phi(s_v)$. 
But for the sake of numerical stability in the training process,
we use 
$\mA_{\phi(s_u),\phi(s_v)}/\max_{\phi(s_i),\phi(s_j)}\mA_{\phi(s_i),\phi(s_j)}$
as a measure. 

Thus the loss function is defined as:

\begin{equation} \label{eq:edloss}
    \mathcal{L}=\sum_{\phi(s_u),\phi(s_v) \in \mathcal{V}} 
    \big[ \mathbf{D}(g(s_u),g(s_v))-\mA_{\phi(s_u), \phi(s_v)}/\max_{\phi(s_i),\phi(s_j)}\mA_{\phi(s_i),\phi(s_j)} \big] ^2.
\end{equation}

This loss function can enforce the subgoal representation provided by the encoder
to respect neighbouring features in the graph. 

Note that in each training phase of the graph encoder-decoder (except the first one), we use the values of the parameters obtained from the last training phase as the initial point, which helps save computation cost.

\subsection{Adaptive training schedule of the graph encoder-decoder}
\label{sec:AT}
The graph stores evolving data, including state representations as node features and connection information in the weighted adjacency matrix $\mA$, which are continuously updated during online training. Since the graph structure and content change at varying rates across episodes, training the graph encoder-decoder at fixed intervals can cause several issues:
(1) high variance in earlier episodes, where sparse or unstable graph data may lead to unreliable model updates;
(2) data underutilization, where intermediate graph states are overwritten before being used for training; and
(3) overfitting in later episodes, as the model repeatedly trains on increasingly redundant data.
To address these issues, we introduce an adaptive training schedule for the graph encoder-decoder, described in the following paragraph.

There are two types of data changing in the graph: node replacement
and edge update. 
We introduce a variable $c$ to track the weighted number of data changes. 
Since the replacement of nodes has a much higher impact on the data than the edge update, we add $N-1$ to $c$ if a node replacement occurs, and add $1$ to $c$ if an edge update happens:
\begin{equation} \label{eq:updateschedule}
  c =
    \begin{cases}
      c + N-1, & \text{if a node replacement happens,}\\
      c + 1, & \text{if an edge update happens.} 
    \end{cases}       
\end{equation}
When this variable exceeds a certain value, specifically a tolerance $\beta$ multiplied by
the total number of non-diagonal elements $N^2-N$ in the matrix $\mA$. 
we perform one gradient update for the graph encoder-decoder,  and
then we reset $c$ to $0$.

\subsection{Hierarchical agent with graph encoder-decoder}
The proposed method involves traditional goal-conditioned settings and a subgoal representation extractor implemented by a graph encoder-decoder.

The high-level policy $\pi_h(g_t|\phi(s_t))$
nominates a subgoal every $K$ steps and is trained using the external environmental reward $r_{\text{ext}}$. 

The policy can be implemented by any policy-based RL algorithm that takes transition tuples $(s_t, g_t, a_t, r_t, s_{t+1}, g_{t+1})$ as input. 
To encourage it to propose a subgoal that is not too difficult to reach from the current state $s_t$ for more efficient exploration, 
we add an intrinsic term to the high-level reward, considering the distance between the subgoal features of $s_t$ and $g_t$ in the subgoal space:
\begin{equation} \label{eq:hir}
r_h(s_t, g_t, s_{t+1})= r_{\text{ext}}+r_{\text{int}}=r_{\text{ext}}+\alpha_h \cdot \mathbf{D}(\mathbf{E}(\phi(s_t)), \mathbf{E}(g_t)),
\end{equation}
where $\alpha_h$ is a hyperparameter that controls the significance
of the intrinsic term in the high-level reward.

The low-level policy $\pi_l(a_t|\phi(s_t), g_t)$, 

however, operates in the subgoal space. While it still takes $\phi(s_t)$ and $g_t$ as input and outputs an atomic action $a_t$, we compute the reward based on distances in the subgoal space:
\begin{equation} \label{eq:lir}
r_l(s_t, g_t, a_t, s_{t+1})=-\|\phi(s_{t+1})-g_t\|^2 + \alpha_l \cdot \mathbf{D}(\mathbf{E}(\phi(s_{t+1})), \mathbf{E}(g_t)),
\end{equation}
where $\alpha_l$ is a hyperparameter controlling the significance of the reward term in the low-level reward. By computing the intrinsic reward in the subgoal space rather than in the state space, the function provides high values when proposed subgoals are easy to reach from the current location and low values when subgoals are close in the original state space but difficult to reach from the current location. The low-level agent can also be any policy-based algorithm.

\subsection{Balancing between speed and performance}\label{ss:balance}
Due to the excessive comparisons between the current state representation and the node features during graph updates,

as well as the training cost of the graph encoder-decoder on a large graph, the experiments show that the GCHRL method, after incorporating the method, takes approximately twice as long as the original GCHRL method.

To reduce the additional cost, we can 
either decrease the frequency of sampling candidates for node features, train the graph encoder-decoder with a subset of all available training data, or do both.

For the sampling frequency, instead of comparing the state representation with node features 
in each time step, we do it in every $t_c$ time steps.
This may significantly speed up the method while maintaining  satisfactory performance.

To reduce the training data in each graph encoder-decoder training cycle, we randomly sample node pairs in the graph instead of using every node pair for training.

In the experiments section, we will present the training time and performance of G4RL with the above two techniques applied.

\subsection{Results}

\paragraph{Environmental settings} We used AntMaze, AntGather, AntPush, AntFall and AntMaze-Sparse environments from the GYM MuJoCo library \citep{todorov2012mujoco}. The first four involve complex navigation and manipulation tasks performed by a simulated multi-armed robot, while AntMaze-Sparse presents a particularly challenging scenario due to sparse reward signals, providing feedback only upon reaching the goal. Note that AntPush and AntFall contain asymmetric transitions which lead to irreversible state changes that can be difficult to learn with an undirected graph. We added these environments to show G4RL's robustness in inherently asymmetric environments. For the state representation, we selected a subset of raw state dimensions that contains spatial information (e.g. coordinates and arm angles) to serve as the node representation for the graph encoder-decoder. We deliberately aligned the choice of environments with prior work to ensure a fair and consistent comparison between the backbone algorithms and their G4RL-augmented versions. This allowed us to use the same environments, hyperparameters, and codebases provided by the original studies. Our goal was to demonstrate that G4RL can consistently enhance the performance of these backbone algorithms under comparable conditions.

\paragraph{Baselines} For baseline methods, we select typical GCHRL methods that use a Euclidean distance as the intermediate reward for the low-level agent to assess G4RL's capability to correct the bias of this reward and improve the performance of these baselines.

\begin{figure}[h!]
    \centering
    \subfigure[]{\includegraphics[width=0.32\textwidth]{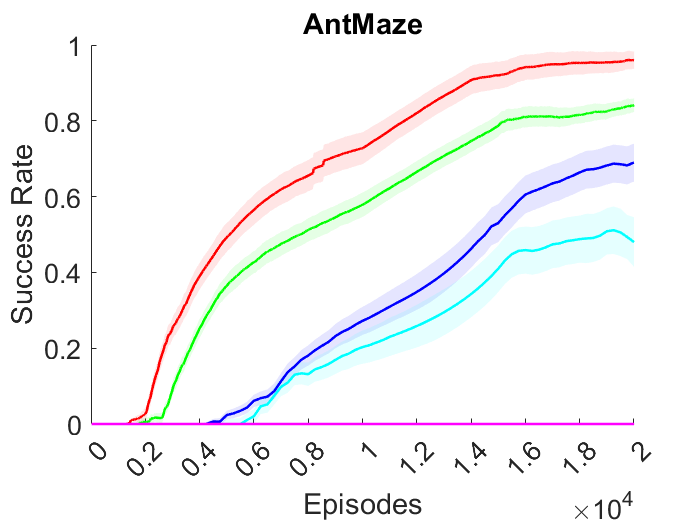}} 
    \subfigure[]{\includegraphics[width=0.32\textwidth]{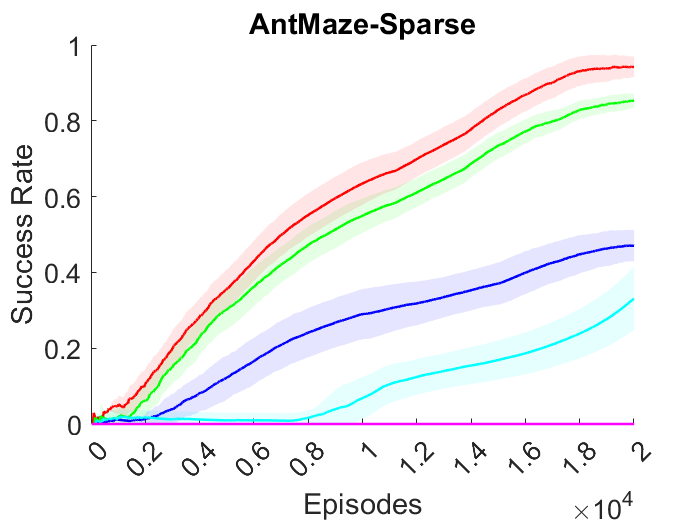}} 
    \subfigure[]{\includegraphics[width=0.32\textwidth]{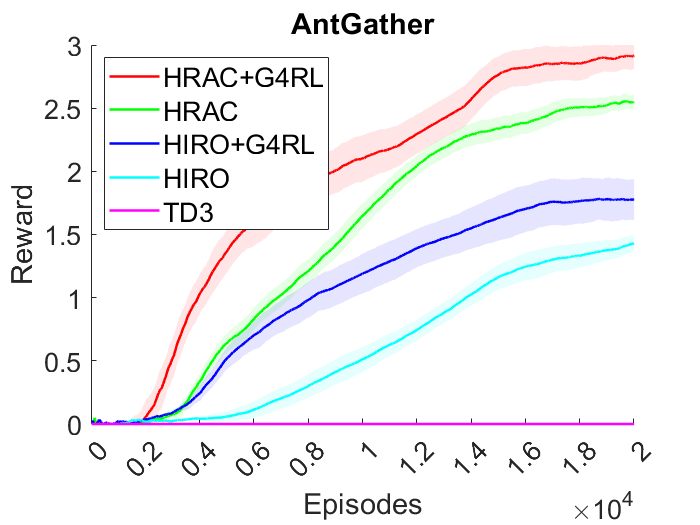}}
    \caption{Success Rate on (a) AntMaze (b) AntMaze-Sparse 
    and Reward on (c) AntGather, using HIRO, HIRO-G4RL, HRAC, HRAC-G4RL, and TD3. Incorporating G4RL in HIRO and HRAC significantly enhances their performance.}
    \label{fig:comp}
\end{figure}

\begin{figure}[h!]
    \centering
    \subfigure[]{\includegraphics[width=0.32\textwidth]{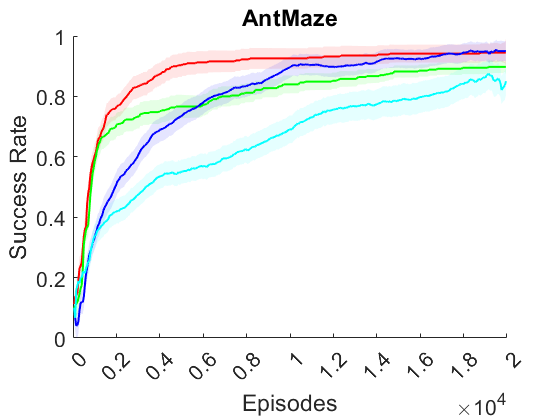}} 
    \subfigure[]{\includegraphics[width=0.32\textwidth]{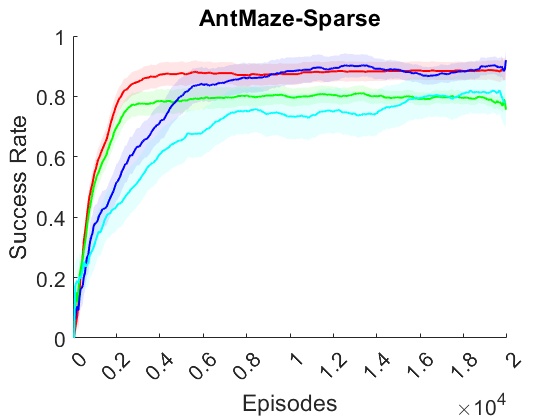}} 
    \subfigure[]{\includegraphics[width=0.32\textwidth]{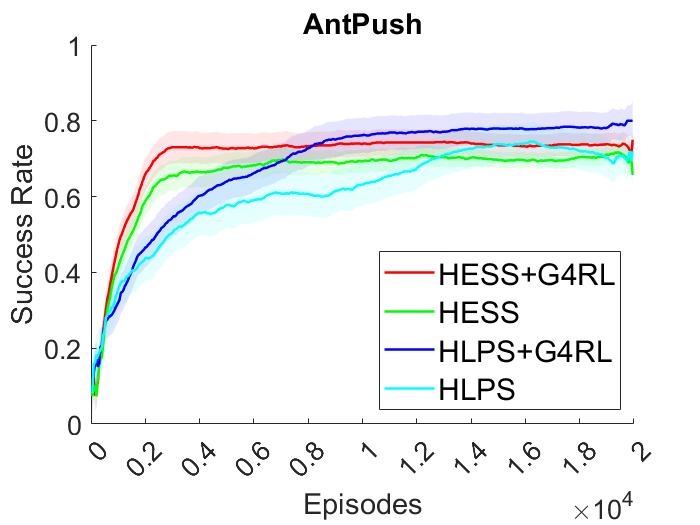}}
    \caption{Success Rate on (a) AntMaze (b) AntMaze-Sparse 
    and (c) AntPush, using HESS, HESS-G4RL, HLPS, HLPS-G4RL. Incorporating G4RL in HESS and HLPS significantly enhances their performance.}
    \label{fig:comp1}
\end{figure}

The learning curves of baseline methods and G4RL-applied versions are plotted in Figure \ref{fig:comp} and \ref{fig:comp1}. Note that all the curves reported are averages from 20 independent runs and they have been equally smoothed for better visualization.

From Figures \ref{fig:comp} and \ref{fig:comp1}, we observe that, in all environments,
incorporating G4RL in the base GCHRL methods significantly enhances their performance, further improving the already strong results of these hierarchical methods 
compared to the non-hierarchical method. 
Notably, G4RL-augmented methods not only achieve higher final success rates but also converge substantially faster, with the most significant improvements observed during the early stages of training.

\begin{figure}[h!]
    \centering
    \subfigure[]{\includegraphics[width=0.32\textwidth]{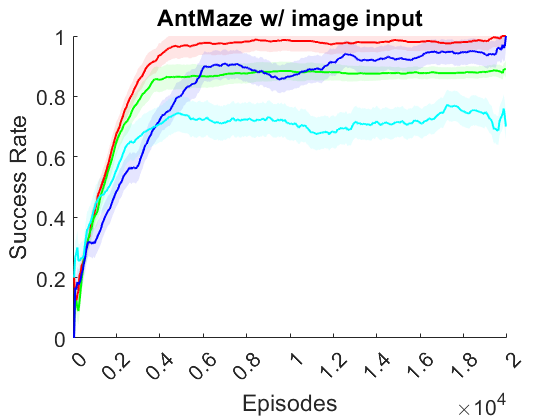}} 
    \subfigure[]{\includegraphics[width=0.32\textwidth]{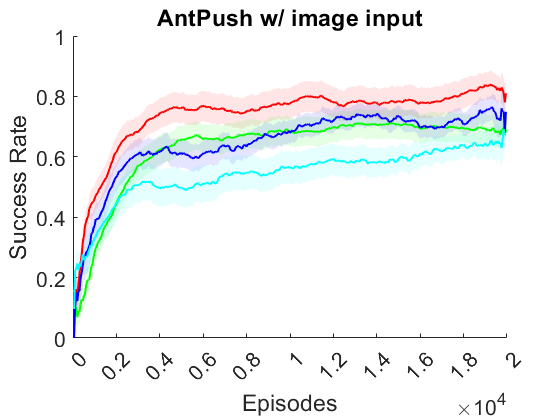}} 
    \subfigure[]{\includegraphics[width=0.32\textwidth]{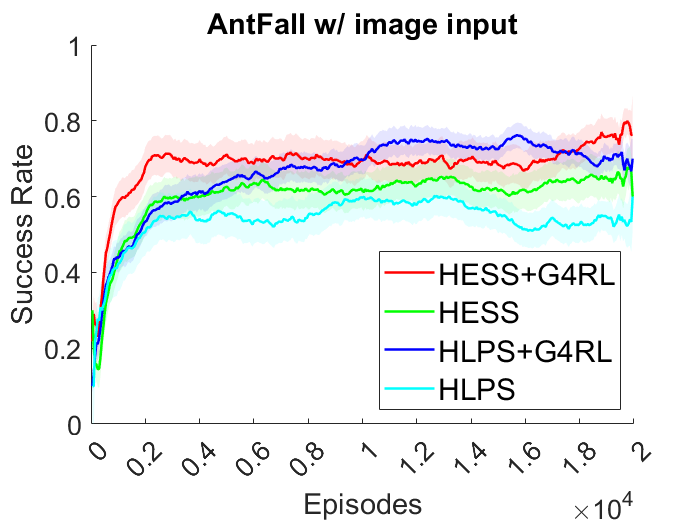}}
    \caption{Success Rate on (a) AntMaze (b) AntPush 
    and (c) AntFall with image state features, using HESS, HESS-G4RL, HLPS, HLPS-G4RL. Incorporating G4RL helps convergence and achieves higher performance across all tested image-based environments.}
    \label{fig:image}
\end{figure}

To demonstrate the proposed method's effectiveness in environments with image-based state representations, we conducted experiments on AntMaze, AntPush, and AntFall, utilizing images as states, and compared the results with HESS and HLPS, along with their G4RL variations. We use Mean Squared Error (MSE) to measure the pixel-wise differences between image states to decide whether a new node should be added to the graph. 
The test results, given in Figure \ref{fig:image}, show that methods incorporating G4RL exhibit faster convergence and achieve higher performance across all tested image-based environments.

\subsubsection{The effect of high/low-level intrinsic reward}
We consider the following variants of G4RL to show the effectiveness of adding high-level and low-level intrinsic rewards:
\begin{itemize}
    \item \textbf{High+Low-level intrinsics}: Apply both equation (\ref{eq:hir}) and equation (\ref{eq:lir}) to the high-level and low-level rewards respectively.
    \item \textbf{High-level intrinsic only}: Apply equation (\ref{eq:hir}) to the high-level rewards and set $\alpha_l=0$ in equation (\ref{eq:lir}) when it is applied 
    to the low-level rewards. 
    \item \textbf{Low-level intrinsic only}: Apply equation (\ref{eq:lir}) to the low-level rewards and 
    set $\alpha_h=0$ in equation (\ref{eq:hir}) when it is applied  to the high-level rewards.
    \item \textbf{HIRO/HRAC/HESS/HLPS}:  Vanilla baseline methods.
\end{itemize}

Same as before, all the curves reported in this section are drawn from results averaged across 20 independent runs. All curves have been equally smoothed for better visualization.

\begin{figure}[h!]
    \centering
    \subfigure[]{\includegraphics[width=0.32\textwidth]{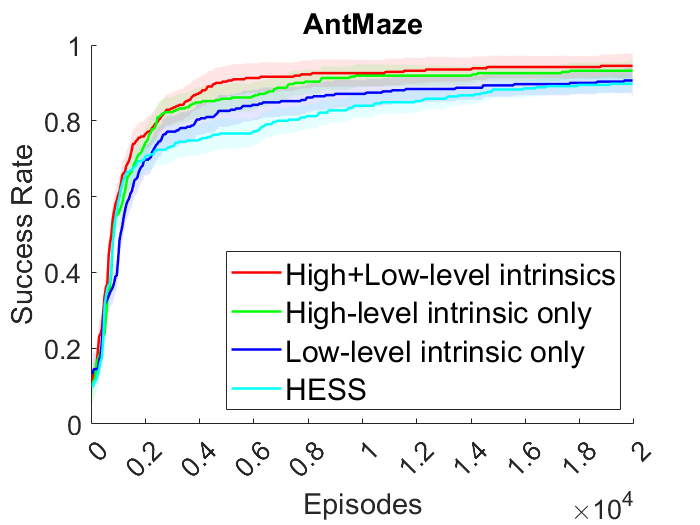}} 
    \subfigure[]{\includegraphics[width=0.32\textwidth]{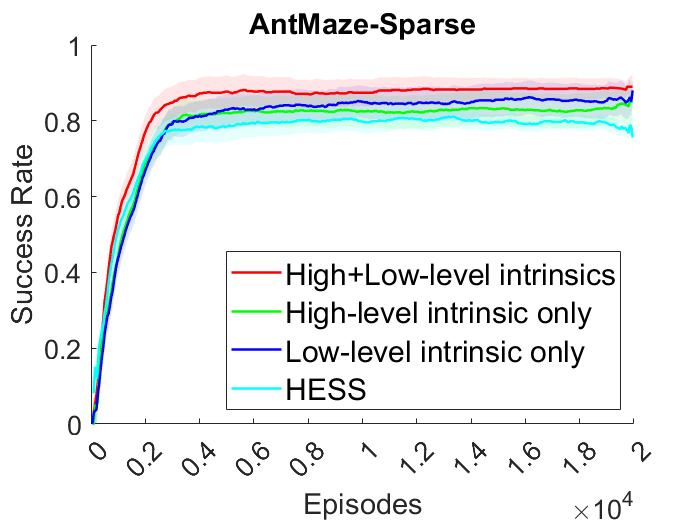}} 
    \subfigure[]{\includegraphics[width=0.32\textwidth]{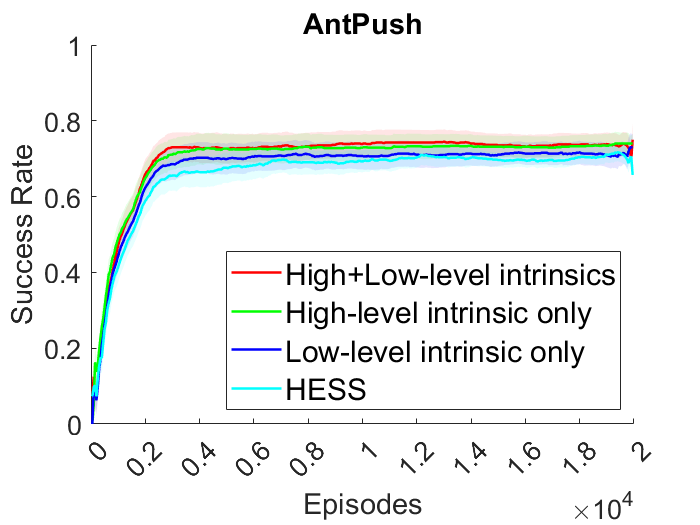}}
    \caption{Success Rate on (a) AntMaze (b) AntMaze-Sparse 
    and (c) AntPush using HESS-G4RL, HESS + High-level intrinsic, HESS + Low-level intrinsic and HESS. All curves have been equally smoothed for better visualization. The combination of high-level and low-level intrinsic rewards results in the highest success rates and fastest convergence.}
    \label{fig:HESS_ab}
\end{figure}

\begin{figure}[h!]
    \centering
    \subfigure[]{\includegraphics[width=0.32\textwidth]{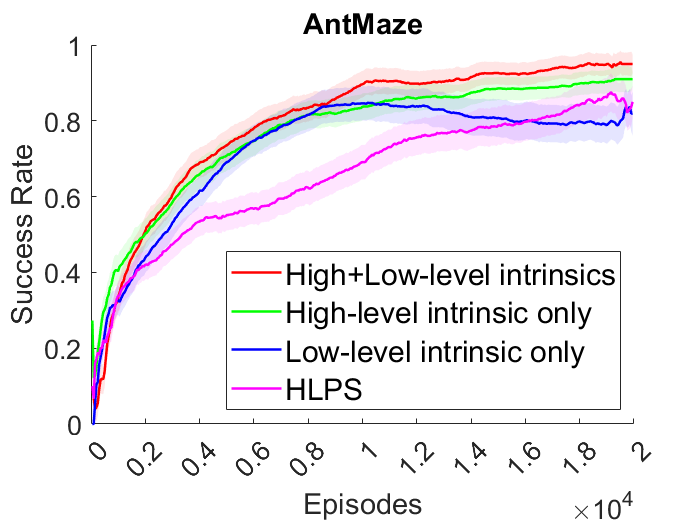}} 
    \subfigure[]{\includegraphics[width=0.32\textwidth]{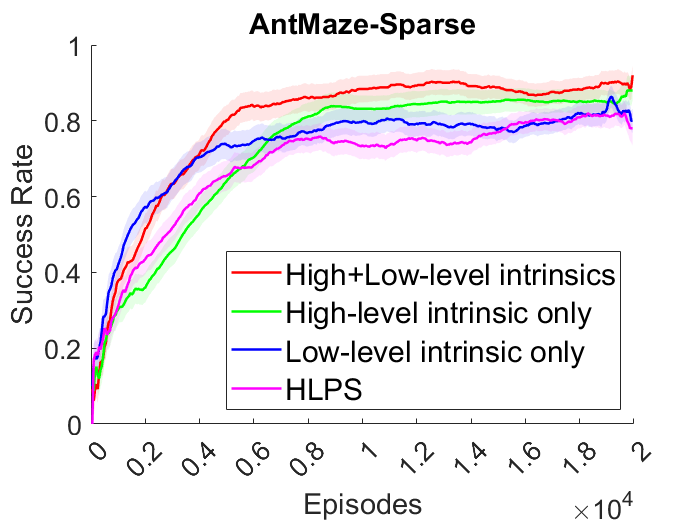}} 
    \subfigure[]{\includegraphics[width=0.32\textwidth]{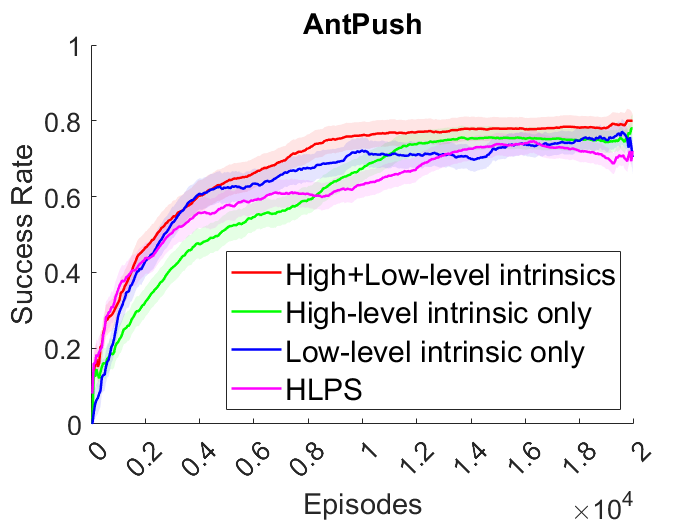}}
    \caption{Success Rate on (a) AntMaze (b) AntMaze-Sparse 
    and (c) AntPush using HLPS-G4RL, HLPS + High-level intrinsic, HLPS + Low-level intrinsic and HLPS. All curves have been equally smoothed for better visualization. The combination of high-level and low-level intrinsic rewards results in the highest success rates and fastest convergence.}
    \label{fig:HLPS_ab}
\end{figure}



Figures \ref{fig:HESS_ab} to \ref{fig:HLPS_ab} show that, across all tested environments and algorithms, the combination of high-level and low-level intrinsic rewards results in the highest success rates and fastest convergence. The high-level intrinsic-only variant outperforms the low-level intrinsic-only variant, especially in sparse reward tasks, indicating that high-level intrinsic rewards play a crucial role in facilitating efficient exploration by encouraging the agent to select reachable and meaningful subgoals. In contrast, low-level intrinsic rewards have limited effect on exploration, primarily refining the execution of local behaviors. These results demonstrate that intrinsic rewards at different hierarchy levels serve complementary functions, and their combination yields superior performance.

\begin{figure}[h!]
    \centering
    \subfigure[]{\includegraphics[width=0.4\textwidth]{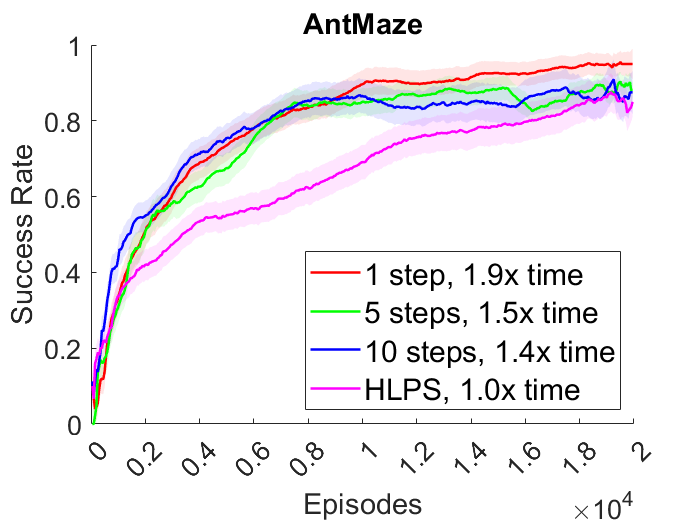}} 
    \subfigure[]{\includegraphics[width=0.4\textwidth]{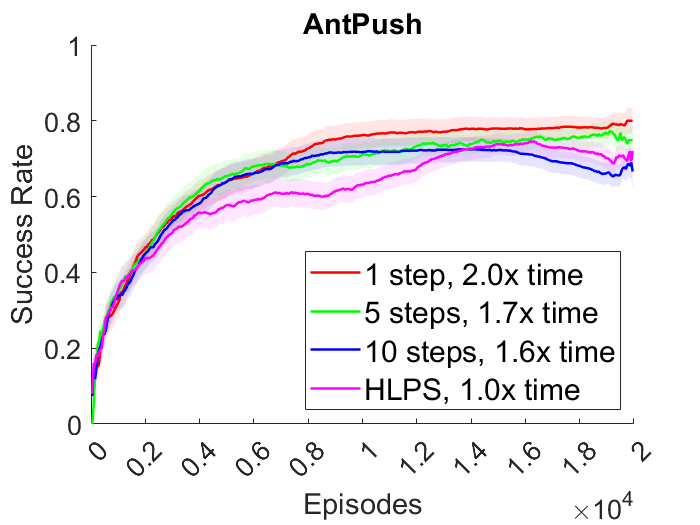}} 
    \caption{Success Rate on (a) AntMaze and (b) AntPush using HLPS+G4RL and HLPS. The number of steps in the legend indicates the selection of $t_c$ as described in Section \ref{ss:balance} and the timescale is calculated w.r.t. the vanila HLPS algorithm. Increasing the sampling interval substantially reduces computation time, with only minor degradation in success rates across both AntMaze and AntPush tasks.}
    \label{fig:HLPS_sample}
\end{figure}

\begin{figure}[h!]
    \centering
    \subfigure[]{\includegraphics[width=0.4\textwidth]{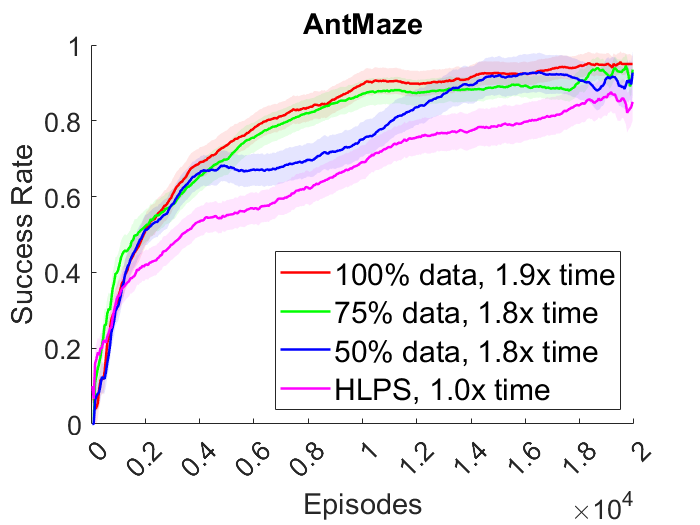}} 
    \subfigure[]{\includegraphics[width=0.4\textwidth]{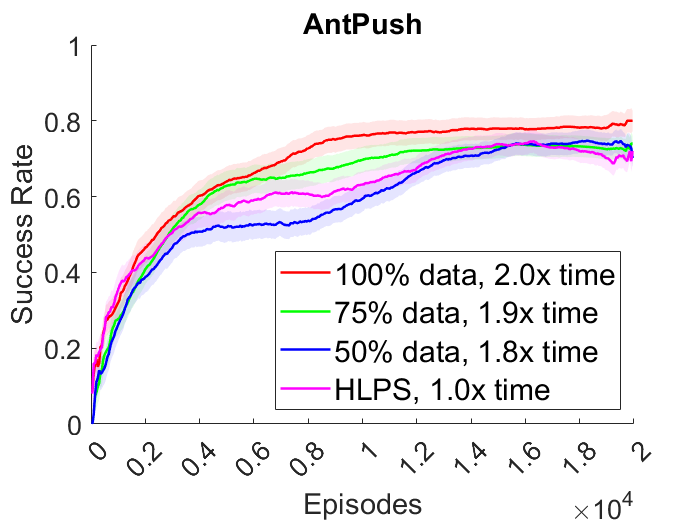}} 
    \caption{Success Rate on (a) AntMaze and (b) AntPush using HLPS+G4RL and HLPS. The percentage of data in the legend indicates the amount of data used in the training of the graph encoder-decoder as described in Section \ref{ss:balance} and the timescale is calculated w.r.t. the vanila HLPS algorithm. Reducing the amount of training data leads to only marginal improvements in computational efficiency and has negligible impact on final performance.}
    \label{fig:HLPS_data}
\end{figure}

To assess the trade-off between computational efficiency and performance, we evaluate two acceleration strategies mentioned in Section \ref{ss:balance}. First, we vary the sampling frequency of node features by testing intervals of 1, 5, and 10 steps in HLPS. As shown in Figure \ref{fig:HLPS_sample}, increasing the sampling interval substantially reduces computation time, as it decreases the number of interactions with the graph, with only minor degradation in success rates across both AntMaze and AntPush tasks. Second, we vary the proportion of training data used for the graph encoder-decoder, testing $50\%$, $75\%$, and $100\%$ subsets. Results in Figure \ref{fig:HLPS_data} indicate that reducing the amount of training data leads to only marginal improvements in computational efficiency and has negligible impact on final performance.

These findings suggest that the primary computational bottleneck of G4RL lies in the graph construction and node comparison processes described in Section \ref{sec:construction}, rather than in the encoder-decoder training itself. Adjusting the sampling frequency is therefore an effective approach for reducing time cost while largely preserving the benefits of G4RL integration.

\subsection{Conclusion}

This work has presented a novel approach using a graph encoder-decoder to address the challenges of poor subgoal representations and sample inefficiency in GCHRL.
The proposed architecture is designed to efficiently evaluate unseen states by operating in the graph representation space. It is easy to implement and can be seamlessly integrated into any existing GCHRL algorithms to enhance their performance in primarily symmetric environments. 
The experiments on both sparse and dense control tasks have demonstrated the effectiveness and robustness of the method.

Despite the advantages demonstrated by G4RL in GCHRL tasks, its effectiveness depends significantly on several hyperparameters (e.g., $\epsilon_d$, $\alpha_l$, and $\alpha_h$), which require careful tuning to achieve optimal performance across different environments. In future work, we aim to develop methods for automatically selecting these hyperparameters based on environmental dynamics, thereby reducing the need for manual tuning. Also, we plan to extend our work by exploring how to generate subgoals with more interpretable representations to facilitate knowledge transfer, potentially leveraging alternative graph representations (e.g. Graph Laplacian). Another promising direction is to transfer the knowledge embedded in the graph structure to new tasks by analyzing graph topology and establishing mappings between nodes of different state graphs. Additionally, more evidence on asymmetric environments is needed to demonstrate G4RL's robustness when asymmetric/irreversible transitions are present.

\section{Shapley Credit Assignment for Denser RLHF}
\label{2ndwork}
This section details our second published work, \methodfullname{} (\methodacronym) \citep{cao2025scar}.
We first review the standard RLHF setup and the inherent reward sparsity problem. We then introduce a game-theoretic perspective on text generation for credit assignment, define \methodacronym{} based on Shapley values, and finally discuss efficient approximation techniques.

\subsection{RLHF and Reward Sparsity}
\label{subsec:rlhf_sparsity}
We formulate the autoregressive text generation process using the standard Markov Decision Process (MDP) formalism, denoted by $\mathcal{M} = (\mathcal{S}, \mathcal{A}, P, R, \gamma)$.
The process begins in an initial state $s_0 \in \mathcal{S}$, which corresponds to the input prompt $x$. At each discrete time step $t \in \{0, 1, \dots, T-1\}$, the agent (the language model) is in state $s_t$, representing the concatenation of the prompt and the sequence generated thus far: $s_t = x \oplus y_{1:t}$ (where $y_{1:0}$ denotes an empty sequence, so $s_0=x$, and $\oplus$ denotes token concatenation). 
The agent selects an action $a_t \in \mathcal{A}$, which corresponds to choosing the next token $y_{t+1}$ from the vocabulary $\mathcal{V}$ according to its policy $\pi_\theta(a_t | s_t)$, parameterized by $\theta$.
We identify the action space with the vocabulary, i.e., $\mathcal{A}=\mathcal{V}$.
The state transition function $P(s_{t+1} | s_t, a_t)$ is deterministic in this context. Upon taking action $a_t=y_{t+1}$ in state $s_t$,
the system transitions to next state $s_{t+1} = s_t \oplus a_t = x \oplus y_{1:t+1}$. This generation process continues until a maximum sequence length $T$ is reached or a designated end-of-sequence (EOS) token is generated. 
For notational simplicity, we often assume a fixed horizon $T$. The discount factor $\gamma$ is typically set to 1 in finite-horizon text generation tasks.

In standard RLHF pipeline \citep{NIPS2017_d5e2c0ad, NEURIPS2020_1f89885d, NEURIPS2022_b1efde53}, a reward model $R_\phi$, parameterized by $\phi$, is trained beforehand on a dataset of human preferences $\mathcal{D}_{\text{pref}} = \{(y^w, y^l)_i\}$, where $y^w$ is preferred over $y^l$ by human annotators. The reward model assigns a scalar score $r_\phi(x, y)$ reflecting the quality or preference level of a \emph{completed} sequence $y$ given the prompt $x$.

To stabilize training and prevent the policy $\pi_\theta$ from drifting too far from a reference distribution (often the initial pre-trained LLM, denoted $\pi_{\text{ref}}$), the reward signal used for optimization is typically augmented with a penalty term at each step $t$. A common choice is the Kullback-Leibler (KL) divergence between the current policy $\pi_\theta$ and the reference policy $\pi_{\text{ref}}$. The standard objective is often formulated as:
\begin{equation}
\label{eq:rlhf_objective}
\mathcal{J}(\theta) = \mathbb{E}_{x \sim \mathcal{D}, y \sim \pi_\theta} \left[ \sum_{t=1}^T R^{\text{orig}}_t \right]
\end{equation}
where $\cal D$ is the dataset and the reward at each timestep $t$ in the standard sparse setup is given by
\begin{equation}
\label{eq:original_reward}
R^{\text{orig}}_t = R^{\text{KL}}_t + \mathbb{I}(t=T) \cdot r_\phi(x, y)
\end{equation}
Here, $R^{\text{KL}}_t = -\beta \log(\pi_\theta(y_t|x, y_{<t}) / \pi_{\text{ref}}(y_t|x, y_{<t}))$ is the KL penalty associated with timestep $t$, $\beta$ is the KL coefficient, and $\mathbb{I}(t=T)$ is an indicator function ensuring the reward model score $r_\phi(x, y)$ is assigned only at the final timestep $T$. This terminal assignment makes the reward signal inherently \emph{sparse}, posing a significant challenge for credit assignment during RL training. Such sparsity directly undermines the efficiency of the learning process and frequently leads to the convergence to suboptimal policies \citep{ng1999policy, NIPS2016_afda3322, NEURIPS2023_b8c90b65}. To overcome this, a principled approach to redistribute the terminal reward more densely across the generation steps is desirable. 


\subsection{A Game-Theoretic Framework for Credit Assignment}
\label{subsec:game_framework}
We frame the generation of a sequence $y$ for a given prompt $x$ as a cooperative game. Let the generated text $y$ be segmented into $N$ contiguous units, $y = (u_1, u_2, \dots, u_N)$. These units could be tokens, spans, sentences, or paragraphs depending on the task. The ``players'' in this game are these $N$ text units. Let $\mathcal{P} = \{u_1, \dots, u_N\}$ be the set of players.

\paragraph{Characteristic Function.}
The value of cooperation among a subset (coalition) $S \subseteq \mathcal{P}$ of players is defined by a characteristic function $v: 2^\mathcal{P} \to \mathbb{R}$. This function should quantify the collective contribution of the units in $S$ towards the final reward objective. We define $v(S)$ based on the reward model's evaluation of the partial text sequence formed by concatenating the units $\{u_i \mid u_i \in S\}$ in their original order. Let $y_S$ denote this concatenated partial sequence. Then, the value function is defined as:
\begin{equation}
\label{eq:characteristic_function}
v(S) = r_\phi(x, y_S)
\end{equation}
For the empty set, $v(\emptyset) = 0$. The value of the grand coalition $v(\mathcal{P})$ corresponds to the original sparse reward for the complete sequence, $v(\mathcal{P}) = r_\phi(x, y)$. 
Note that evaluating $r_\phi$ on partial sequences $y_S$  requires careful consideration, as $y_S$ represents an incomplete sequence. Ideally, $v(S)$ could represent the expected reward obtained by keeping the units in $S$ fixed and sampling the remaining units from the current policy $\pi_\theta$. In the implementation, we resort to a practical approximation. To evaluate $v(S)$, we construct a sequence in which the tokens belonging to units $u_i \in S$ are placed in their original order. The positions corresponding to units $u_j \notin S$ are filled using empty spaces.

\paragraph{Shapley Value Calculation.}
The Shapley Value $\mathrm{SV}_{u_i}(v)$ for a player $u_i \in \mathcal{P}$ (text unit $u_i$) quantifies its fair contribution to the grand coalition value $v(\mathcal{P})$, calculated as the average marginal contribution of player $u_i$ over all possible permutations of player arrivals:
\begin{equation}
\label{eq:shapley_value}
\mathrm{SV}_{u_i}(v) = \sum_{S \subseteq \mathcal{P} \setminus \{u_i\}} \frac{|S|! \, (N - |S| - 1)!}{N!} [v(S \cup \{u_i\}) - v(S)].
\end{equation}
The Shapley values uniquely satisfies axioms such as \textit{efficiency} ($ \sum_{u_i \in \mathcal{P}} \mathrm{SV}_{u_i}(v) = v(\mathcal{P}) $), \textit{symmetry} (equal reward for equal contribution), \textit{linearity}, and the \textit{null player} property (no contribution means no credit), making it a principled choice for fair credit allocation \citep{shapley1953stochastic}.

\subsection{Shapley Values as Dense Rewards} 
We use the calculated Shapley values to define a dense reward signal for the RL agent. Let unit $u_i$ consist of tokens generated from timestep $t_{i-1}+1$ up to and including timestep $t_i$ (with $t_0=0$, so $t_i$ marks the completion timestep of unit $u_i$). We assign the Shapley value $\mathrm{SV}_{u_i}(v)$ associated with unit $u_i$ as an additional reward component specifically at the timestep $t_i$ marking the completion of that unit. Let $R^{\text{shap}}_t$ denote this Shapley-based reward at timestep $t$. Then,
\begin{equation}
\label{eq:srs_reward_time}
R^{\text{shap}}_t = \begin{cases} \mathrm{SV}_{u_i}(v) & \text{if } t = t_i \text{ for some unit } u_i \\ 0 & \text{otherwise} \end{cases}
\end{equation}
This component distributes the total reward $r_\phi(x,y)$ across the episode based on the Shapley contributions, since $\sum_{t=1}^T R^{\text{shap}}_t = \sum_{i=1}^N \mathrm{SV}_{u_i}(v) = r_\phi(x,y)$ (due to the efficiency property, where $N$ is the total number of units).

We then define the total reward $R_t$ provided to the RL agent at timestep $t$ as a convex combination of the dense Shapley-based credit allocation and the original sparse terminal reward, while retaining the per-step KL penalty. Using a hyperparameter $\alpha \in [0, 1]$, the total reward is:
\begin{equation}
\label{eq:total_dense_reward_convex}
R_t(\alpha) = R^{\text{KL}}_t + \alpha \cdot R^{\text{shap}}_t + (1 - \alpha) \cdot \mathbb{I}(t=T) \cdot r_\phi(x, y)
\end{equation}
Here, $\alpha$ controls the interpolation:
\begin{itemize}
    \item If $\alpha = 0$, $R_t(0) = R^{\text{orig}}_t$, recovering the standard sparse reward signal.
    \item If $\alpha = 1$, $R_t(1) = R^{\text{KL}}_t + R^{\text{shap}}_t$. The terminal reward $r_\phi(x,y)$ is entirely replaced by an equivalent value distributed densely according to Shapley contributions throughout the episode.
    \item If $0 < \alpha < 1$, the agent receives both the intermediate Shapley-based rewards and a residual portion of the original terminal reward.
\end{itemize}
This formulation allows flexible control over the density of the reward signal, balancing immediate feedback with the final outcome signal.

\begin{theorem}[Policy Invariance under \methodacronym{} Reward Shaping]
    Consider a parameterized language model $\pi_\theta$ with a learned reward model $R_\phi$. Let $\mathcal{M} = (\mathcal{S}, \mathcal{A}, P, R^{\text{orig}}_t, \gamma)$ be the original MDP with its reward from the reward model and $\widehat{\mathcal{M}} = (\mathcal{S}, \mathcal{A}, P, R_t(\alpha), \gamma)$ be the MDP with dense Shapley reward. If $\pi_\theta$ is optimal for $\widehat{\mathcal{M}}$, then $\pi_\theta$ is also optimal for $\mathcal{M}$, and vice versa.
\end{theorem}

\subsection{Efficient Approximation of Shapley Values}
\label{sec:shap_approx}

The direct calculation of Shapley values using Equation~\eqref{eq:shapley_value} necessitates evaluating the characteristic function $v(S)$ (defined in Eq.~\eqref{eq:characteristic_function}) for all $2^N$ possible coalitions $S$ of the $N$ text units. This exponential complexity renders exact computation practically infeasible for typical sequence lengths encountered in text generation \cite{shapley1953stochastic}. To make \methodacronym{} practical, we employ two key strategies: adaptive segmentation of text into units  and efficient approximation of their Shapley values.

\paragraph{Adaptive Text Segmentation as Players.}
The definition of ``players'' (text units $u_i$) in the cooperative game is crucial for both interpretability and computational tractability. We adapt the granularity of these units based on the task and the typical length of the generated responses, aiming to keep the number of players $N$ manageable. We experiment with three levels of segmentation:
\begin{itemize}
    \item \textbf{Token-level:} For tasks producing very short responses, each token $y_t$ can be treated as an individual player $u_i$. This offers the finest granularity but results in a larger $N$.
    \item \textbf{Span-level:} For medium-length responses, we leverage constituency parsing \citep{marcus-etal-1993-building} to establish a hierarchical grammatical structure over the generated sequence $y$. This process yields a constituency tree where tokens (leaf nodes) are organized into hierarchically nested constituents. Players are then defined as these syntactic constituents (e.g., noun phrases, verb phrases), formed by grouping tokens that share a common parent or ancestor node within this tree. This approach reduces $N$ while preserving semantic coherence within each player unit, as constituents are inherently meaningful linguistic units.\footnote{\url{https://www.nltk.org/howto/parse.html}}
    \item \textbf{Sentence-level:} For tasks generating longer, multi-sentence responses, each sentence in the output $y$ constitutes a player. Segmentation is achieved using standard sentence boundary detection. This approach markedly reduces $N$, especially for verbose outputs.
\end{itemize}
The choice of segmentation strategy is a hyperparameter, allowing a trade-off between the granularity of credit assignment and the computational cost of Shapley value estimation.

\paragraph{Approximation Using Owen Values.}
To ensure the practical applicability of \methodacronym{}, we employ an approximation scheme based on Owen value \citep{owen1977values}, which is a coalitional extension of Shapley values designed for games where players are grouped into a predefined coalition structure\citep{aumann1974cooperative}.
For the task, a hierarchical structure $\mathcal{B}$ is imposed on the sequence of $N$ text units, achieved by applying a heuristic parsing algorithm to the units. This partition $\mathcal{B}$ defines nested groupings of the units. The Owen value is then computed with respect to this partition $\mathcal{B}$.

Let $N$ be the set of all players, and let $v: 2^N \to \mathbb{R}$ be the characteristic function.
The \textbf{coalition structure} is a partition of $N$, denoted by $\mathcal{B} = \{B_1, B_2, \ldots, B_m\}$, where $m = |\mathcal{B}|$ is the number of unions.
For any player $i \in N$, let $B(i)$ be the unique union in $\mathcal{B}$ that contains player $i$. The Owen value $\Phi_i(v, \mathcal{B})$ for player $i$ is given by the formula:

\begin{equation}
    \Phi_i(v, \mathcal{B}) = \sum_{R \subseteq \mathcal{B} \setminus \{B(i)\}}  \frac{|R|! (m - |R| - 1)!}{m!}  \cdot \text{MC}
\end{equation}

Where $\text{MC}$ is the marginal contribution, defined as:

\begin{equation}
    \text{MC} = v \left( \bigcup_{B \in R} B  \cup \{B(i)\} \right) - v \left( \bigcup_{B \in R} B  \right)
\end{equation}

Marginal contributions are evaluated by forming coalitions structurally: combinations involving subsets within a unit's own group are explored, while units belonging to other groups in the partition are treated as indivisible blocks, as they are either entirely included or entirely excluded from a coalition, rather than exploring all their individual subsets.
By limiting the evaluation to coalitions dictated by the partition structure $\mathcal{B}$, the number of required characteristic function evaluations (reward model queries) is substantially reduced compared to the exact Shapley computation. Consequently, the computational complexity is reduced from exponential, $O(2^N)$, to quadratic in $N$, rendering the approach tractable. We use the SHAP package \cite{NIPS2017_7062} for Shapley values and Owen values computation.

\subsection{Results}

In this section, we empirically evaluate the effectiveness of \methodacronym{} across three distinct tasks characterized by varying response lengths. My primary objective is to demonstrate that \methodacronym{} enables more efficient and effective training compared to standard sparse RLHF and alternative dense reward baselines.

\begin{figure}[h]
\begin{small}
\begin{tabular}{@{}lp{\dimexpr0.915\textwidth-2\tabcolsep\relax}@{}} 
\toprule
\textbf{Prompt:} & ``While some scenes were'' \\
\midrule[1pt] 

\textbf{Sparse:} &
  \plainscore{initially}{0.0}\hspace{0.4em}%
  \plainscore{disturbing}{0.0}\hspace{0.4em}%
  \plainscore{to}{0.0}\hspace{0.4em}%
  \plainscore{sit}{0.0}\hspace{0.4em}%
  \plainscore{through,}{0.0}\hspace{0.4em}%
  \plainscore{they}{0.0}\hspace{0.4em}%
  \plainscore{ultimately}{0.0}\hspace{0.4em}%
  \plainscore{contributed}{0.0}\hspace{0.4em}%
  \plainscore{to}{0.0}\hspace{0.4em}%
  \plainscore{a}{0.0}\hspace{0.4em}%
  \plainscore{deeply}{0.0}\hspace{0.4em}%
  \plainscore{powerful}{0.0}\hspace{0.4em}%
  \plainscore{and}{0.0}\hspace{0.4em}%
  \plainscore{moving}{0.0}\hspace{0.4em}%
  \plainscore{story}{0.0}\hspace{0.4em}%
  \plainscore{.}{0.0}\hspace{0.4em}%
  \plainscore{\texttt{<EOS>}}{+10.0} \\
\midrule

\textbf{ABC:} &
  \hlscore{initially}{0.0}{poscontrib!1!white}\hspace{0.4em}%
  \hlscore{disturbing}{0.0}{poscontrib!1!white}\hspace{0.4em}%
  \hlscore{to}{0.0}{poscontrib!1!white}\hspace{0.4em}%
  \hlscore{sit}{0.0}{poscontrib!1!white}\hspace{0.4em}%
  \hlscore{through,}{0.0}{poscontrib!1!white}\hspace{0.4em}%
  \hlscore{they}{0.0}{poscontrib!1!white}\hspace{0.4em}%
  \hlscore{ultimately}{0.0}{poscontrib!1!white}\hspace{0.4em}%
  \hlscore{contributed}{0.0}{poscontrib!1!white}\hspace{0.4em}%
  \hlscore{to}{0.0}{poscontrib!1!white}\hspace{0.4em}%
  \hlscore{a}{0.0}{poscontrib!1!white}\hspace{0.4em}%
  \hlscore{deeply}{+0.8}{poscontrib!18!white}\hspace{0.4em}%
  \hlscore{powerful}{+0.8}{poscontrib!18!white}\hspace{0.4em}%
  \hlscore{and}{+0.2}{poscontrib!12!white}\hspace{0.4em}%
  \hlscore{moving}{+1.3}{poscontrib!23!white}\hspace{0.4em}%
  \hlscore{story}{+1.6}{poscontrib!26!white}\hspace{0.4em}%
  \hlscore{.}{+5.3}{poscontrib!63!white}\hspace{0.4em}%
  \hlscore{\texttt{<EOS>}}{0.0}{zerocontrib!15!white} \\
\midrule

\textbf{SCAR:} &
  \hlscore{initially}{+0.8}{poscontrib!18!white}\hspace{0.4em}%
  \hlscore{disturbing}{-0.5}{negcontrib!15!white}\hspace{0.4em}%
  \hlscore{to}{-0.1}{negcontrib!11!white}\hspace{0.4em}%
  \hlscore{sit}{-1.8}{negcontrib!28!white}\hspace{0.4em}%
  \hlscore{through,}{-0.8}{negcontrib!18!white}\hspace{0.4em}%
  \hlscore{they}{-0.2}{negcontrib!12!white}\hspace{0.4em}%
  \hlscore{ultimately}{+1.0}{poscontrib!11!white}\hspace{0.4em}%
  \hlscore{contributed}{+0.7}{poscontrib!17!white}\hspace{0.4em}%
  \hlscore{to}{0.0}{poscontrib!1!white}\hspace{0.4em}%
  \hlscore{a}{+0.5}{poscontrib!15!white}\hspace{0.4em}%
  \hlscore{deeply}{+1.8}{poscontrib!28!white}\hspace{0.4em}%
  \hlscore{powerful}{+2.2}{poscontrib!32!white}\hspace{0.4em}%
  \hlscore{and}{+0.5}{poscontrib!15!white}\hspace{0.4em}%
  \hlscore{moving}{+4.0}{poscontrib!50!white}\hspace{0.4em}%
  \hlscore{story}{+0.7}{poscontrib!17!white}\hspace{0.4em}%
  \hlscore{.}{+0.8}{poscontrib!18!white}\hspace{0.4em}%
  \hlscore{\texttt{<EOS>}}{0.0}{zerocontrib!15!white} \\
 
\bottomrule
\end{tabular}
\end{small}
\caption{Comparison of reward distribution strategies for an example generated sequence. Sparse RLHF assigns the total reward at the end. \methodacronym{} and ABC distribute this reward across tokens/spans based on their respective methodologies, shown with background highlights (color hue for sign, intensity for magnitude; more intense/saturated means higher absolute contribution) and numerical scores.}
\label{fig:reward_distribution_comparison}
\end{figure}

\paragraph{Evaluation Tasks and Models.}
We evaluate the proposed method across three diverse tasks prevalent in RLHF research: sentiment control, text summarization, and instruction tuning \citep{chan2024dense,  yoon-etal-2024-tlcr}. For sentiment control and instruction tuning, we utilize the implementation in \citep{chan2024dense}. However, for summarization, due to difficulties in reproducing the results, we switched to the implementation in \citep{huang2024the}. The datasets, policy models, and reward models used for each task are described in more detail below.

\textbf{Sentiment Control:} The objective is to generate positive reviews of movies. We use the IMDB dataset \citep{maas-etal-2011-learning}. The policy model is GPT-2 small \citep{radford2019language}, initialized by fine-tuning for one epoch on the IMDB training set. During RLHF training, prompts are generated by randomly selecting the first 4 to 8 tokens from reviews in the training data. The reward signals are provided using a pre-trained sentiment classifier, same as \citep{chan2024dense}.

\textbf{Text Summarization:}
We evaluate the method on the automatic text summarization task, following prior work \citep{NEURIPS2020_1f89885d, chan2024dense, lee2023rlaif}. For this evaluation, we use the Reddit TL;DR dataset \citep{volske-etal-2017-tl}, specifically the filtered version \citep{NEURIPS2020_1f89885d}, which includes approximately 116K training examples, 6K validation examples, and 6K test examples. The policy model used is Pythia-1B \citep{biderman2023pythia}, which we initialize via supervised fine-tuning on the training set for 2,000 steps with a batch size of 64. Additionally, we train a 1B-parameter reward model (initialized using the SFT model) on 92K human preference pairs, achieving approximately 74\% accuracy on the validation set.

\textbf{Instruction Turning:} We evaluate models on the task of following user instructions. To do this, we fine-tune language models using the helpfulness subset of the Anthropic Helpful and Harmless (HH) dataset \citep{bai2022training}, which contains 43K human-written prompts paired with model responses that have been ranked by human annotators for helpfulness. Preference is based on which response is more informative and helpful for the task. The policy model is initialized using the OpenLLaMA-7B model \citep{openlm2023openllama}, an open-source reproduction of Meta's LLaMA collection \citep{touvron2023llama} trained on fully open-source dataset. For the reward model, we use a 3B reward model \citep{dong2023raft}. This reward model was trained using the same HH dataset, where it learns to assign a scores to candidate completions based on their predicted usefulness.

\paragraph{Baselines.} We compare the proposed method against three key baselines. \textbf{RLHF} represents the standard approach using the sparse terminal reward (Eq.~\ref{eq:original_reward}) with KL regularization. \textbf{ABC (Attention Based Credit)} \citep{chan2024dense} uses reward model attention scores for dense rewards distribution. \textbf{Uniform} is a baseline that distributes the terminal reward evenly across all tokens. 
For fair comparison, all methods are optimized using the PPO objective \citep{schulman2017proximal} with consistent hyperparameters. All methods initialize their policy models using the same SFT checkpoint to ensure a common starting point. All experiments were conducted on a single A100 GPU (80GB VRAM), and results are averaged over $5$ random seeds.


\paragraph{Evaluation Metrics.} We track the average reward $r_\phi(x, y)$ per episode during training to evaluate learning speed and the level of convergence. The final performance is reported as the mean reward on the test set after convergence. For the summarization task, we additionally employ \textbf{LLM-as-a-judge} \citep{zheng2023judging} evaluation to compare the quality (e.g., accuracy, coverage, conciseness, clarity, and coherence) of summaries generated by models trained with different methods. We randomly sampled 1K summaries from the TL;DR test set for LLM evaluation. To mitigate potential positional bias in these pairwise comparisons, we randomize the presentation order of summaries. For the instruction-tuning task, we use AlpacaEval \citep{alpaca_eval} to compare the quality of 1K model's response. AlpacaEval is designed to better handle potential issues such as length bias, thereby providing a more reliable assessment of response quality.

Figure~\ref{fig:reward_distribution_comparison} provides a qualitative illustration of how \methodacronym{} distributes rewards compared to sparse RLHF and ABC for an example generated sequence. Sparse RLHF, by definition, assigns the entire reward only at the end of the sequence. ABC, which uses attention scores from the reward model's final layer to distribute rewards, tends to concentrate rewards on tokens near the end of the sequence. As seen in the example, significant credit is assigned to the final punctuation mark (``.''), while earlier, crucial tokens receive almost zero attention scores. Furthermore, standard attention scores are non-negative, making it difficult for ABC to assign explicit negative credit to tokens or spans that detract from the output quality. For instance, a phrase like ``\textit{disturbing to sit through}'' that negatively impact the perceived sentiment, would not receive negative rewards from ABC. In contrast, \methodacronym{} can assign both positive and negative rewards to tokens based on their game-theoretic marginal contribution.

\begin{figure}[h!]
    \centering
    \subfigure[]{\includegraphics[width=0.32\textwidth]{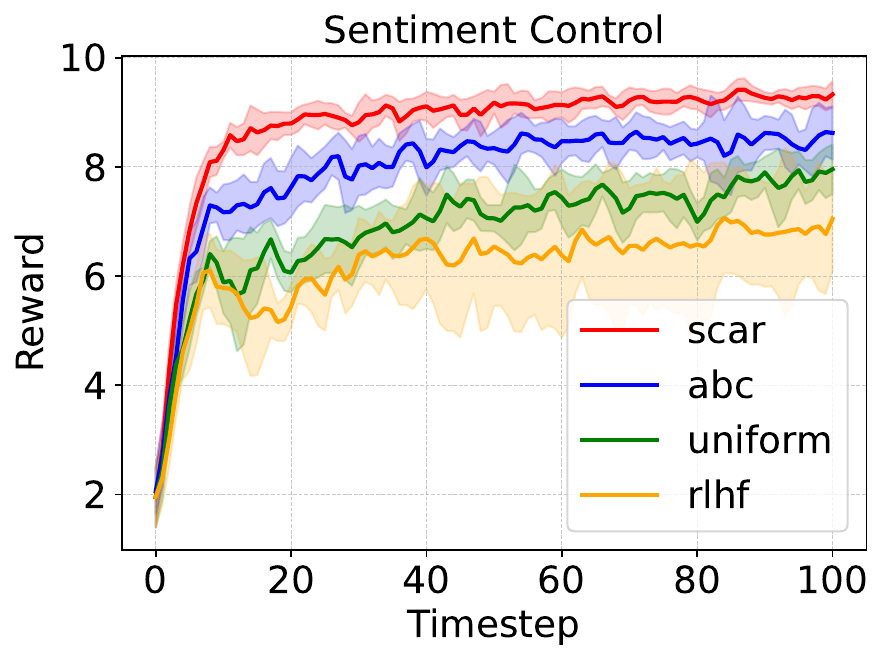}} 
    \subfigure[]{\includegraphics[width=0.32\textwidth]{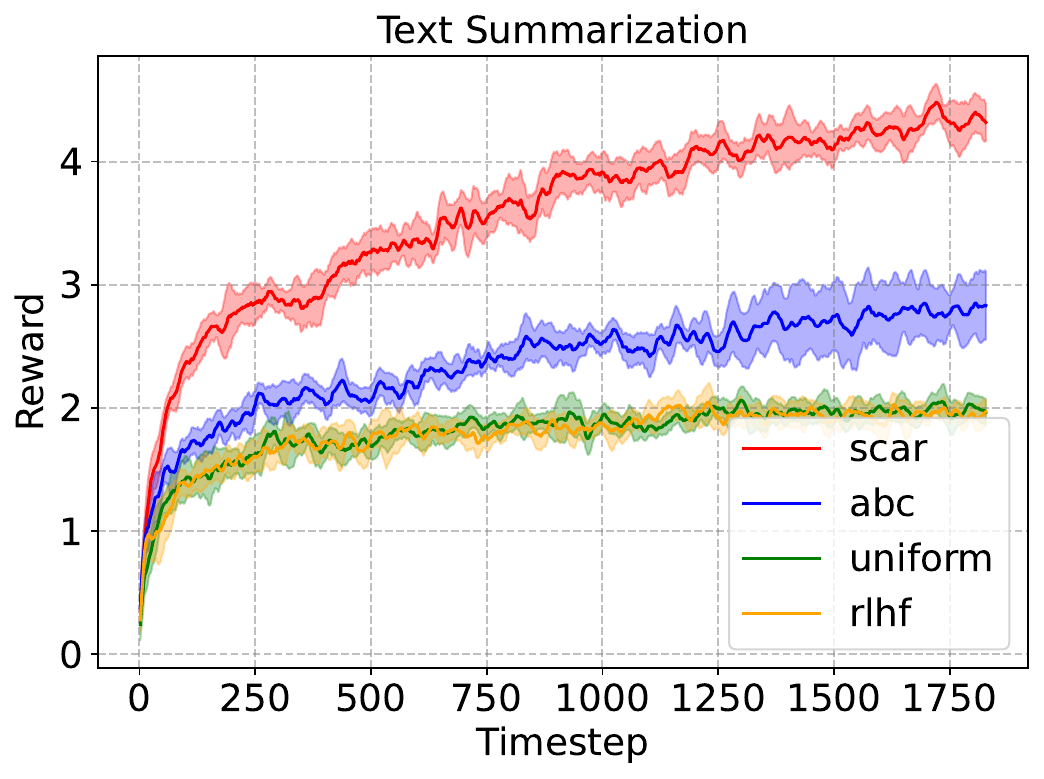}}
    \subfigure[]{\includegraphics[width=0.32\textwidth]{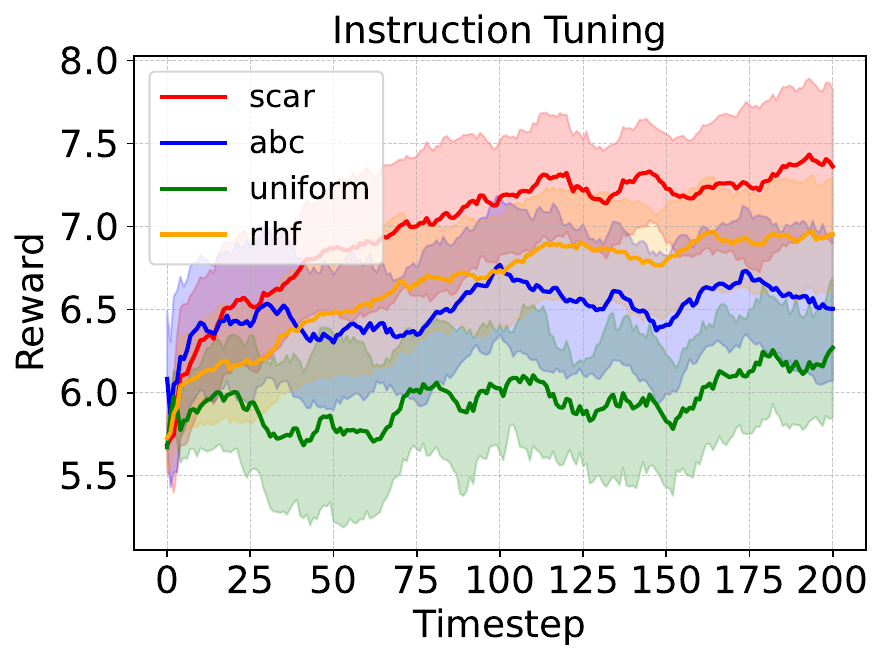}}
    \caption{Average reward per timestep during RLHF training for sentiment control (left), text summarization (center), and instruction tuning (right). Curves show the mean reward across five random seeds, with shaded regions representing the standard deviation. \methodacronym{} consistently demonstrates faster convergence and achieves higher or comparable final reward levels compared to sparse RLHF, Uniform reward distribution, and Attention-Based Credit (ABC) baselines.}
    \label{fig:reward_curves}
\end{figure}


\begin{table*}[htbp]
\centering
\begin{adjustbox}{width=0.75\textwidth, center} 
\begin{tabular}{lccccc}
\toprule
Task & Sparse RLHF & Uniform & ABC & \methodacronym{}  \\
\midrule
IMDB & 6.86$\pm$ 0.86 & 7.73 $\pm$ 0.02 & 8.48 $\pm$ 1.60 & \textbf{9.27 $\pm$ 0.00} \\
TL;DR & 1.60 $\pm$ 0.11 & 1.68 $\pm$ 0.02 & 2.85 $\pm$ 0.21 & \textbf{4.35 $\pm$ 0.11} \\
HH-RLHF & 6.93 $\pm$ 0.00 & 6.17 $\pm$ 0.00 & 6.59 $\pm$ 0.01 & \textbf{7.31 $\pm$ 0.01} \\
\bottomrule
\end{tabular}
\end{adjustbox} 
\caption{Average reward scores for the trained policy on the test sets for sentiment control (IMDB), text summarization (TL;DR), and instruction tuning (Anthropic HH). Higher scores indicate better performance. Results are averaged over 5 random seeds. Best performance per task is in \textbf{bold}.}
\label{tab:main_results}
\end{table*}

\begin{wraptable}{r}{0.45\textwidth} 
\centering
\footnotesize
\begin{tabular}{@{}lcc@{}} 
\toprule
~ & Baselines & Win (\%) \\
\midrule
\multicolumn{3}{@{}l}{\textit{Text Summarization (Reddit TL;DR)}} \\ 
& vs. RLHF & 61.2\% \\
& vs. ABC & 60.3\% \\
\midrule
\multicolumn{3}{@{}l}{\textit{Instruction Tuning (Anthropic HH)}} \\ 
& vs. RLHF & 56.3\% \\
& vs. ABC & 54.9\% \\
\bottomrule
\end{tabular}
\caption{LLM-as-Judge pairwise win rates for \methodacronym{} against baselines.}
\label{tab:llm_judge_merged}
\end{wraptable}
\newpage
As depicted in Figure~\ref{fig:reward_curves}, \methodacronym{} consistently demonstrated advantages over three baseline methods in terms of learning speed and convergence across all three tasks.
This consistent pattern across diverse tasks suggests that the principled, Shapley value-based credit assignment offered by the method effectively improves learning efficiency and enhances the final policy performance. Table~\ref{tab:main_results} presents the average reward scores achieved by each method on the held-out test sets for the three tasks. As shown in the table, \methodacronym{}-tuned policy consistently achieves the highest performance compared to the baselines.

For summarization, we use \texttt{gemini-2.5-pro} to compare anonymized model outputs (\methodacronym{} vs. baselines) on coherence, relevance, conciseness, and overall quality.
As shown in Table~\ref{tab:llm_judge_merged}, summaries generated by the \methodacronym{}-tuned model were preferred over those from the ABC-tuned model in 60.3\% and over the sparse RLHFC-tuned model in 61.2\%.
For the instruction tuning task, we leveraged AlpacaEval \citep{alpaca_eval} and used \texttt{gpt-4-turbo} as an LLM judge, to conduct robust pairwise evaluations. These evaluations assessed helpfulness, harmlessness, and adherence to instructions. \methodacronym{}-generated responses achieved a win rate of 54.9\% when compared against ABC, and a win rate of 56.3\% against standard sparse RLHF. These results provide further evidence that the improvements from \methodacronym{} translate to genuinely higher-quality outputs according to human-like preferences.
\subsection{Conclusion}
This part has presented a novel approach using a graph encoder-decoder to address the challenges of poor subgoal representations and sample inefficiency in GCHRL.
The proposed architecture is designed to efficiently evaluate unseen states by operating in the graph representation space. It is easy to implement and can be seamlessly integrated into any existing GCHRL algorithms to enhance their performance in primarily symmetric environments. The experiments on both sparse and dense control tasks have demonstrated the effectiveness and robustness of the method.

Despite its strengths, \methodacronym{} has limitations: the computational overhead of Shapley approximations, even with optimizations like Owen values and adaptive segmentation; the assumption that the reward model can meaningfully score partial sequences, which may not suit certain types like rule-based models that only evaluate final answers (e.g., in mathematical reasoning).
Future work will target more efficient approximation techniques, robust and adaptive segmentation methods, and rigorous evaluation on larger-scale language models and broader tasks.
\section{Proposed Research: Dense reward from previous knowledge for continual RL}
\label{3rdwork}
The ability to continually learn and adapt to new environments \cite{khetarpal2022towards} while retaining knowledge of previously visited ones is crucial for modern RL algorithms. Previous methods focus on mitigating plasticity loss in neural networks \cite{abbas2023loss, dohare2024loss, dohare2023maintaining}. Recently, people have tried to develop RL agents which can acquire structured knowledge and adapt to new environments \cite{anand2023prediction}. This requires that the agent can 1) store the knowledge acquired in previous environments in a structured, transferable form and 2) adapt the stored knowledge to the new environment/domain easily. When the agent learns in tasks with similar dynamics, this similarity in task structure can be considered as "invariant knowledge" and can be used to transfer. To achieve this, we naturally consider graphs as the carriers of transferable knowledge, which are suitable for modelling environmental dynamics. The approximated state graphs model the environmental dynamics in terms of node connectivity and degrees. Furthermore, the information in the graph nodes/labels is easy to transfer to the new environment.

\paragraph{Graph construction in the first environment:} 
The graph construction process is similar to what we discussed in Section \ref{sec:construction}.
Suppose the representation is different from any state representations stored in the graph. In that case, we store the state representation $\phi(s_t)$ as the node feature of an empty node in the graph and build an edge between this node and the node that corresponds to the previous state. In addition, we store the estimated current state value into the node label:
\begin{equation} 
\forall_{s_v \in \mathcal{V}}, \,  \|\phi(s_t)-\phi(s_v)\|_2>\epsilon_d,
\end{equation}
\begin{equation} 
\mA_{\phi(s_t), \phi(s_{t-1})}=\mA_{\phi(s_{t-1}), \phi(s_t)}=1,\ \ \vy_{s_t}=V_\pi(s_t),
\end{equation}
where $\vy_{s_t}$ represents the state value of $s_t$.

Graph updating in the first environment: Suppose the graph is now full.
When a new state $s_t$ is encountered, 

\begin{equation} 
s_v = \argmin_{s_u: \|\phi(s_t)-\phi(s_u)\|_2 \leq \epsilon_d} \|\phi(s_t)-\phi(s_u)\|_2.
\end{equation}
If $s_v$ exists, as before we relabel the node as $\phi(s_t)$ and evaluate 
the state value:
\begin{equation}
    \vy_{s_t}=V_\pi(s_t)
\end{equation}
Otherwise, we replace the oldest state node in the graph with the current state node,
delete all edges previously linked to that node,
and create an edge $(s_{t-1},s_t)$ with weight $\A_{\phi(s_{t-1}),\phi(s_t)}=\A_{\phi(s_t),\phi(s_{t-1})}=1$ 
and set 
$\vy_{s_t}=V_\pi(s_t)$.

After sufficient learning in the first environment $E_1$, we calculate the graph laplacian of the state graph $\mL$ 
\begin{equation}
    \mL = \mD - \mA
\end{equation}
where $\mD$ is the diagonal degree matrix. 
For later uses, this $\mL$ will be denoted by $\mL_1$.

\paragraph{Comparing with knowledge from the first environment in the second environment} 
We introduce a second environment $E_2$ where the task is designed structurally analogous to the first, yet varies in its specifics, requiring retraining within the traditional RL framework. For this environment, we construct and update a graph using the identical procedure. At regular intervals, we find the graph Laplacian $\mL_2$. While canonical labelling \cite{babai1983canonical} is the standard tool for verifying exact graph isomorphism, it is ineffective when graphs are only approximately isomorphic. Therefore, we employ a spectral method to quantitatively assess the degree of similarity by measuring the difference between the associated Laplacian matrices $\mL_1$ and $\mL_2$.
We compute the eigen decompositions: 
\begin{equation}
    \mL_k = \mV_k \mathbf{\Lambda}_k \mV^T_k
    = \lambda_k^{(1)}\vv_k^{(1)} (\vv_k^{(1)})^T 
      + \lambda_k^{(2)}\vv_k^{(2)} (\vv_k^{(2)})^T
      + \ldots 
      + \lambda_k^{(N)}\vv_k^{(N)} (\vv_k^{(N)})^T, \ \ k=1,2,
\end{equation}
where $\mV_k=[\vv_k^{(1)}, \vv_k^{(2)}, \ldots,\vv_k^{(N)}]$ is orthogonal 
and the largest entry of each eigenvector $\vv_k^{(i)}$ in magnitude is positive,
and $\mathbf{\Lambda}_k=\diag(\lambda_k^{(i)})$
with $\lambda_k^{(1)}\geq \lambda_k^{(2)} \geq \cdots \geq \lambda_k^{(N)}=0 $.
Each row of $\mV_k$ is a feature vector of the corresponding node of the graph.


Note that the two graphs are isomorphic if and only if there exists 
a permutation matrix $\mP$ such that $\mL_2=\mP \mL_1 \mP^T$.
To check if the two graphs are isomorphic, we first check if the difference between the two spectra is within a preset threshold:
\begin{equation}
    \sum_{j=1}^N |\mathbf{\lambda}_{1}^{(j)}-\mathbf{\lambda}_2^{(j)}|^2 \leq \epsilon_\lambda
\end{equation}
If it holds, we try to match feature vectors of nodes in the two graphs. 
There are two cases:
\begin{enumerate}
    \item All eigenvalues are distinct. 
    In this case, the eigenvectors $\vv_k^{(i)}$
    are unique up to the sign. 
    We try to match rows of $\mV_1$ with rows of $\mV_2$. 
    Specifically, for the first row vector of $\mV_1$, we check if there exists a row vector of $\mV_2$ that is close to it:
    \begin{equation}
        \exists \mV_2(i,:)\; \mbox{ such that } \; \|\mV_1(1,:)-\mV_2(i,:)\|_2\leq\epsilon_v,
    \end{equation}
    where $\epsilon_v$ is a predefined tolerance threshold.
    If the match succeeds, we continue to try to 
    match other rows of $\mV_1$ and $\mV_2$.
    If one match fails, we conclude that the two graphs are not similar.
    
    \item Some eigenvalues are repeated.
    This case is much more complicated. 
    However, there exist  $O(N^3)$ practical methods to do the match,
    see, e.g., \citep{fan2020spectral,klus2018spectral}
 \end{enumerate}
Once two graphs are considered similar, when we act $a_t$ in environment 2 and transition from $s_t$ to $s_{t+1}$ happens, we find the closest state feature to $\phi(s_{t+1})$ stored in graph 2, denoted by $\phi(s_{e2})$ which is the state representation of $s_{e2}$, then find its pairing node in graph 1 (denoted by $s_{e1}$). The intrinsic reward of this transition is then defined as:
\begin{equation}
    R_{int}(s_t,a_t,s_{t+1}) = \beta \vy_{s_{e1}} = \beta V_\pi(s_{e1})
\end{equation}
where $\beta$ is a hyperparameter controlling the significance of the intrinsic term.

\paragraph{Ongoing Work}
To complete this research project, several key areas require further investigation and development:

\begin{itemize}
\item \textbf{Identifying appropriate test environments:} The proposed method is designed to identify and leverage similarities between tasks. To effectively evaluate this capability, it is crucial to select or construct environments that exhibit meaningful task similarities alongside non-trivial differences. For example, we are currently experimenting with variants of the same environment that differ in granularity, or that share the same state space while varying the action space. Careful environment design will be essential to demonstrate the generality and robustness of our approach.

\item \textbf{Establishing relevant baselines:} A rigorous comparison with existing continual reinforcement learning (CRL) methods is necessary to validate our approach. We plan to benchmark against recent baselines such as those proposed by \citep{anand2023prediction} and \citep{abbas2023loss}, using our curated sets of related tasks. Selecting strong and diverse baselines will help clarify where our method stands in terms of performance and generalization.

\item \textbf{Defining evaluation metrics:} In addition to standard CRL performance metrics (e.g., average return or task score), it may be useful to develop task similarity metrics or diagnostics tailored to our method. These could provide deeper insights into when and why our approach succeeds or fails, and guide future improvements.
\end{itemize}


\bibliographystyle{plain} 
\bibliography{ref} 
\end{document}